\documentclass[runningheads]{llncs}
\usepackage[T1]{fontenc}
\usepackage{graphicx}
\usepackage{booktabs}
\usepackage[misc]{ifsym}
\newcommand{\corr}{(\Letter)}
\usepackage{mwe}

\usepackage{amsmath}
\usepackage{amsthm}
\usepackage{amsfonts}
\usepackage{colortbl}
\usepackage{multirow}
\usepackage{adjustbox}
\usepackage{array}

\usepackage{xcolor}
\usepackage{tcolorbox}
\usepackage{fancyhdr}

\fancypagestyle{firstpagefooter}{
  \fancyhf{}  
  \fancyfoot[C]{\small Accepted for publication at ML4CCE workshop at ECML PKDD 2024.}
}

\definecolor{lightyellow}{RGB}{255,255,204}
\definecolor{lightgreen}{RGB}{235,255,235}
\definecolor{lightred}{RGB}{255,204,203}
\definecolor{lightblue}{RGB}{202,225,255} 
\definecolor{validitygreen}{RGB}{0,255,0}
\definecolor{lightgrey}{RGB}{245,245,245} 

\begin{document}

\title{Towards Human-Level Understanding of Complex Process Engineering Schematics: A Pedagogical, Introspective Multi-Agent Framework for Open-Domain Question Answering}

\titlerunning{Bridging the Gap: Towards Human-like Understanding of Process Diagrams}

\author{Sagar Srinivas Sakhinana\inst{1}\corr \and
Geethan Sannidhi\inst{2}  \and
Venkataramana Runkana\inst{1}}

\authorrunning{Sakhinana, S. S., et al.}
\institute{TCS Research, India \email{\{sagar.sakhinana, venkat.runkana\}@tcs.com}
\and
IIIT Pune, India \email{geethansannidhi20@cse.iiitp.ac.in}}
\tocauthor{Sagar Srinivas Sakhinana, Geethan Sannidhi, Venkataramana Runkana}

\maketitle              

\thispagestyle{firstpagefooter}

\vspace{-9mm}
\begin{abstract} 
In the chemical and process industries, Process Flow Diagrams (PFDs) and Piping and Instrumentation Diagrams (P\&IDs) are critical for design, construction, and maintenance. Recent advancements in Generative AI, such as Large Multimodal Models (LMMs) like GPT-4 (Omni), have shown promise in understanding and interpreting process diagrams for Visual Question Answering (VQA). However, proprietary models pose data privacy risks, and their computational complexity prevents knowledge editing for domain-specific customization on consumer hardware. To overcome these challenges, we propose a secure, on-premises enterprise solution using a hierarchical, multi-agent Retrieval-Augmented Generation (RAG) framework for open-domain question answering (ODQA) tasks, offering enhanced data privacy, explainability, and cost-effectiveness. Our novel multi-agent framework employs introspective and specialized sub-agents using open-source, small-scale multimodal models with the ReAct (Reason+Act) prompting technique for PFD and P\&ID analysis, integrating multiple information sources to provide accurate and contextually relevant answers. Our approach, supported by iterative self-correction, aims to deliver superior performance in ODQA tasks. We conducted rigorous experimental studies, and the empirical results validated the proposed approach's effectiveness.

\vspace{-3mm}
\vspace{-8mm}
\end{abstract}

\vspace{-9mm}
\section{Introduction}
\begin{figure*}[!ht]
\centering
\resizebox{0.75\linewidth}{!}{ 
\hspace*{0mm}\includegraphics[keepaspectratio,trim=0.0cm 0cm 0cm 3.0cm,clip]{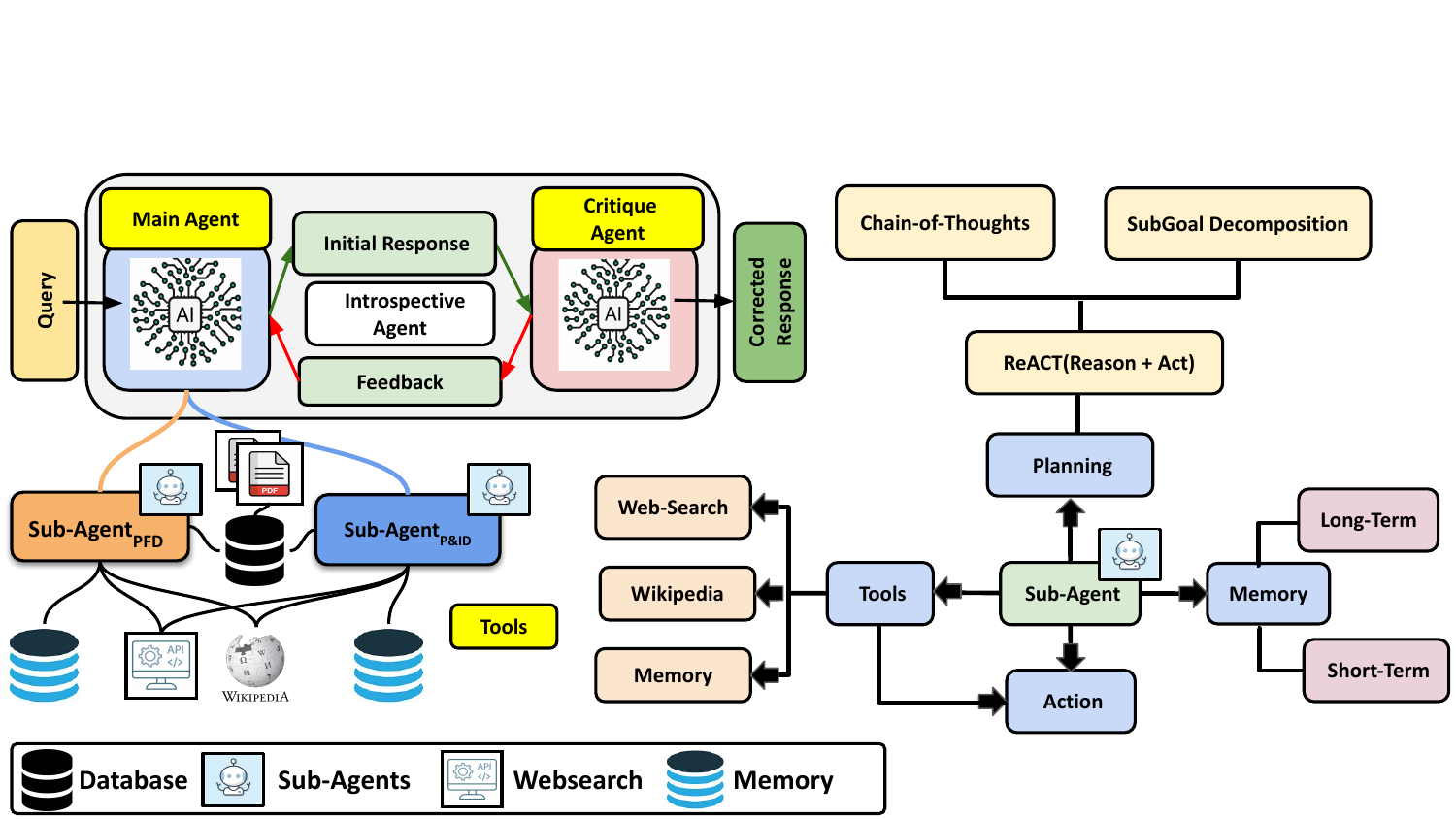} 
}
\vspace{-4mm}
\caption{The figure shows the multi-agent framework for ODQA on complex documents for PFD and P\&ID analysis. It consists of an introspective agent, including a main agent and a critique agent. The main agent orchestrates specialized sub-agents, routing the end-user request to the relevant sub-agent. Each sub-agent utilizes SMMs with the ReACT technique in a four-stage workflow to utilize external tools, allowing dynamic access to specialized resources beyond its pre-trained knowledge: task planning, tool selection, tool calling, and response generation. In task planning, user intent is analyzed and queries are decomposed into sub-tasks. Tool selection involves SMMs selecting appropriate tools (APIs, databases, external knowledge repositories) to solve these sub-tasks. Tool calling involves SMMs extracting the required parameters from the user query and calling the selected tools to retrieve relevant information from document databases, memory databases, web searches, and Wikipedia articles. Memory consists of long-term memory, which stores reusable information for future queries, and short-term memory, which holds session-specific data for immediate processing. Finally, response generation integrates the outputs from these tools with the SMMs' internal knowledge to create comprehensive and coherent responses, and the critique agent iteratively refines the outputs using reflection and correction cycles. We fine-tune SMMs to select appropriate tools and use them accurately during task-specific adaptation to provide accurate and contextually relevant responses. 
}
\label{fig:figure1}
\vspace{-10mm}
\end{figure*}

\vspace{-4mm}
PFDs and P\&IDs play crucial roles in the chemical and process industries, finding applications in various sectors such as oil and gas, pharmaceuticals, the semiconductor industry, and more. PFDs illustrate major equipment interconnections and material/energy flow in a chemical and process plant, while P\&IDs detail piping, instrumentation, and control systems. Both PFDs and P\&IDs are essential documents for the design, construction, operation, and maintenance of chemical and process plants. Recent advancements in Generative AI, such as Large Multimodal Models (LMMs) with advanced vision-language processing capabilities, including OpenAI GPT-4 (Omni)\cite{openai2023gpt4} and Google Gemini\cite{team2023gemini}, have the ability to understand and interpret PFDs and P\&IDs for visual question-answering (VQA). However, using proprietary vision-language models raises data privacy concerns, as the risk of sharing intellectual property could compromise enterprise technological portfolios. Furthermore, the large size and complexity of closed-source LMMs limit their customizability for specialized tasks and knowledge editing. On the other hand, open small-scale multimodal models (SMMs), like Google PaliGemma\cite{team2024gemma} and Microsoft Phi-3, offer the benefits of domain-specific customization and interpretable PFD and P\&ID analysis, but they might fall short in terms of reasoning and generalization capabilities compared to proprietary large-scale models. Developing a secure, on-premises, customizable adaptation of SMMs for PFD and P\&ID analysis provides enterprises with advantages such as enhanced data privacy, explainability, and cost-effectiveness. However, this approach is not without its challenges. In recent times, Retrieval-Augmented Generation (RAG) has combined the strengths of pre-trained SMMs with information retrieval from external knowledge bases for open-ended VQA tasks. However, while RAG techniques allow SMMs to access external databases for VQA, they lack pre-trained knowledge on PFD and P\&ID analysis. Fine-tuning offers task-specific adaptation for PFD and P\&ID analysis but often ignores external databases during VQA, leading to less grounded and reliable answers. To overcome these challenges, we utilize both instruction-tuning and human preference alignment of open-source SMMs to optimally adapt SMMs to domain-specific RAG, addressing the limitations of limited pre-trained domain knowledge and the inability to utilize relevant external knowledge, leading to improved factual accuracy. Customizing SMMs for PFD and P\&ID analysis for VQA tasks, including image captioning and text recognition (OCR), faces two main challenges. First, high-quality human-annotated datasets specific to this domain are scarce. Second, manually annotating PFDs and P\&IDs to generate relevant question-answer (QA) pairs for customizing SMMs is a resource-intensive and time-consuming process that requires expert knowledge and specialized tools. To address the scarcity of human-annotated instruction-tuning datasets, we utilize teacher-student transfer learning (knowledge distillation), where a large model, such as GPT-4, serves as a robust `teacher' to generate instruction-tuning data (image-question-answer pairs) and preference-tuning data (image-question-chosen-rejected pairs) for customizing a `student' – a small-scale model such as PaliGemma – through parameter-efficient fine-tuning (PEFT) methods. This approach enhances grounded language generation and visual reasoning capabilities in task-specific applications, such as image captioning, VQA, and text detection for PFD and P\&ID analysis. In recent years, there has been a surge of interest in advanced RAG applications based on autonomous agents tackling complex goals. These agents utilize vision-language models to achieve several key functionalities: (a) interpreting end-user requests, reasoning about higher-level goals, and breaking down multi-step tasks into simpler, manageable subtasks through task planning; (b) selecting tools, which involves choosing pre-built or custom tools for each subtask and then calling those tools (e.g., external APIs or helper functions) to complete the subtasks; (c) generating responses by synthesizing the information obtained from the tools to create comprehensive and coherent answers; and (d) refining plans using a verify-then-correct approach to analyze and reason about their tool selection choices and usage. Iterative refinement allows for self-evaluation, incorporation of external feedback, and repeated cycles of improvement, leading to more accurate answers. This, coupled with memory augmentation, enables them to maintain context and leverage past conversations to provide coherent answers for multi-turn conversations. More recently, an advanced multi-agent RAG architecture based on autonomous agents has enabled the achievement of high-level goals through enhanced inter-agent communication and collaborative planning. Unlike previous single-agent architectures, this approach can tackle complex, multifaceted tasks with minimal human intervention. In this study, we present a novel hierarchical, multi-agent framework for open-domain question answering (ODQA) in the analysis of complex engineering PFD and P\&ID schematics. The framework consists of an introspective (or meta) agent composed of a main agent and a critique agent. The main agent orchestrates specialized sub-agents and utilizes language models like Google's Gemma with the ReACT technique\cite{yao2022react} for reasoning and decision-making. It interprets complex user queries, delegates tasks to appropriate sub-agents, and forwards the sub-agents' responses to the critique agent. The critique agent evaluates the responses using a Gold LMM (e.g., GPT-4 Turbo) and provides feedback. The framework employs an iterative self-correction process through reflection, incorporating a verify-then-correct process, where the critique agent's feedback is used to iteratively refine the sub-agents' output for improved factual correctness and overall trustworthiness of the framework. Figure \ref{fig:figure1} illustrates the framework.

\vspace{-10mm}

\vspace{0mm}
Our approach begins by extracting text and images from complex PDFs and segmenting the documents using a sliding window technique for granular information retrieval. The text segments are embedded into vector representations and indexed for efficient similarity search. We generate summary descriptions for images and index them. Each specialized sub-agent handles specific diagrams—PFDs and P\&IDs, respectively—and assists the main agent by providing expert analysis in their fields. The sub-agents utilize SMMs with the ReACT technique\cite{yao2022react} to understand and interpret tasks. They employ a retrieve-and-read paradigm, involving tool selection and tool calling to retrieve relevant information from parsed documents, memory databases (to retain and recall previous information), and external APIs for web search or Wikipedia articles. This is followed by reranking to select the most relevant passages. Sub-agents generate multiple answer candidates and corresponding supporting summaries using conditional summarization. The most plausible answer is selected based on the validity and informativeness of its supporting summary, utilizing a two-pronged evaluation process (instance-wise validity and pairwise ranking) to identify the best answer. We evaluate our approach on image captioning, VQA, and text detection (OCR) tasks for PFD and P\&ID analysis. Our findings demonstrate that the framework consistently performs on par with state-of-the-art methods, offering customizability, interpretability, data privacy, and cost-effectiveness.

\vspace{-5mm}
\section{Proposed Method}
\vspace{-4mm}
ODQA addresses a wider range of questions than traditional QA, often relying on vast, unstructured databases or large document collections (e.g., web search, Wikipedia). In this work, we utilize two distinct documents: (a) PFDs (\(D_{PFDs}\)) and (b) P\&IDs (\(D_{P\&IDs}\)). We begin by performing document parsing to extract text and images embedded within complex, unstructured PDFs. Each document is then split into smaller segments using the sliding window chunking technique to preserve context and improve retrieval. A fixed-size window (in words) is moved by a predefined stride to create overlapping chunks. After chunking, each text segment is converted into a vector representation using embedding techniques. These vector representations are indexed in a vector database, allowing for efficient similarity search and retrieval of relevant text chunks based on semantic similarity to user queries. We use GPT-4(omni) to generate alternative text descriptions for the extracted images, which serve as metadata providing content, context, and details of the images. By indexing this metadata, we enable the retrieval of images based on their content, thereby enhancing the effectiveness of multi-modal search queries. Our approach employs a multi-agent framework where the introspective (meta) agent comprises a main agent ($\mathcal{A}^{M}$) and a critique agent ($\mathcal{A}^{C}$). The main agent orchestrates specialized sub-agents ($\mathcal{A}_{PFDs}$, $\mathcal{A}_{P\&IDs}$), each responsible for handling tasks related to PFDs and P\&IDs analysis. The introspective agent utilizes a reflective agentic pattern, consisting of the main agent generating the initial response to user queries and the critique agent performing reflection and correction to evaluate and improve the response. The main agent leverages language models (LMs) like Google Gemma to understand and interpret complex user queries, utilizing the ReACT technique \cite{yao2022react} to enable reasoning and decision-making proficiency. This allows the main agent to strategically approach user queries by breaking them down into smaller, manageable sub-tasks for task delegation to specialized sub-agents and relaying the sub-agent responses to the critique agent. The critique agent utilizes a `Gold LMM-as-a-judge' such as GPT-4 Turbo to evaluate the initial response and provide feedback based on factual correctness and question-answer relevance to improve the response. The main agent routes the sub-task to the specific domain expert sub-agent, where each sub-agent (\(\mathcal{A}_{PFDs}\), \(\mathcal{A}_{P\&IDs}\)) utilizes SMMs, such as PaliGemma. These sub-agents employ the ReACT technique \cite{yao2022react} to analyze and understand the sub-task-specific requirements and objectives through deliberate reasoning and a structured approach to problem-solving. This facilitates strategic tool selection and effective tool use for comprehensive information retrieval and synthesis, leading to more accurate and contextually appropriate outputs. Sub-agents dynamically select tools from the tool inventory based on the context and relevance to the user query, including (a) vector search on both (i) structured memory databases storing previous question-answer pairs and (ii) parsed document storage databases designed to store and retrieve indexed complex PDFs. Other external tools, such as (b) search engines and (c) Wikipedia, are used to retrieve relevant information from the internet or static knowledge bases. By extracting and integrating information from these various sources, the sub-agents provide comprehensive and accurate answers to complex, open-ended questions. To answer a question \(q\), the main agent delegates the user query to the relevant sub-agent. The sub-agent uses a two-step approach known as the retrieve-and-read paradigm. The embedding-based retriever searches a vast collection of information sources, including memory databases, parsed document storage databases, web search results, and Wikipedia articles. The corresponding retrieved passages are denoted as \(C = \{c_{1}, \cdots, c_{M}\}\). The probability of a passage \(c \in C\) being relevant to a given question \(q\) is determined by a softmax function over the similarity scores, expressed as follows:

\vspace{0mm}
\resizebox{0.925\linewidth}{!}{
\hspace{0mm}\begin{minipage}{\linewidth}
\begin{equation}
P(c \mid q) = \frac{\exp(\text{sim}(e_q, e_c))}{\sum_{c' \in C} \exp(\text{sim}(e_q, e_{c'}))} \nonumber
\end{equation}
\end{minipage}
}

where \(c\) represents a passage in the set of passages \(C\). \(e_q\) and \(e_c\) are the embeddings of the question and the passage, respectively. \(\text{sim}(e_q, e_c)\) is a similarity function (e.g., dot product) between the embeddings. We identify a relevant top-\(N\) subset \(C^{+} = \{c_{1}, \cdots, c_{N}\}\) from this larger corpus to maximize recall as follows:

\vspace{-5mm}
\resizebox{0.95\linewidth}{!}{
\hspace{0mm}\begin{minipage}{\linewidth}
\begin{equation}
\quad C^{+} = \operatorname{Retriever}(q, C, N) \nonumber
\end{equation}
\end{minipage}
}

\vspace{0mm}
where \(N\) signifies the number of retrieved passages. After retrieval, the reranker evaluates the relevance of each passage in \(C^{+}\) and selects the top-\(\mathcal{K}\) passages that are most likely to contain the answer, as follows:

\vspace{-1mm}
\resizebox{0.95\linewidth}{!}{
\hspace{0mm}\begin{minipage}{\linewidth}
\begin{equation}
C^{++} = \operatorname{Reranker}(q, C^{+}, \mathcal{K}) \nonumber
\end{equation}
\end{minipage}
}

where \(C^{++}\) is the refined subset of passages, and \(\mathcal{K} \leq N\) is the number of passages forwarded to the reader (sub-agent). The sub-agent `reads' the retrieved information to find and synthesize the necessary details to generate a diverse set of potential answer candidates using a specially designed prompt consisting of the question \(q\) and the retrieved passages \(C^{++}\). It then performs conditional summarization to create high-level summaries of these passages, tailored to provide supporting evidence for each specific answer candidate within the provided context. This approach enhances the sub-agent's ability to process the retrieved passages \(C^{++}\) by emphasizing evidence and logical reasoning, thereby providing targeted summaries that support each specific answer candidate. Given a question \(q\), the retrieved passages \(C^{++}\), and a smaller multimodal model (SMM) \(\mathcal{M}\), we generate \(K\) answer candidates \(\widehat{\mathbf{a}} = [\widehat{a}_{1}, \ldots, \widehat{a}_{K}]\) using a custom prompt \(p_{\text{can}}\), leveraging \(C^{++}\) and \(q\) to guide \(\mathcal{M}\) in generating the answer candidates. Mathematically, this procedure can be represented as:

\vspace{-0mm}
\resizebox{0.95\linewidth}{!}{
\hspace{0mm}\begin{minipage}{\linewidth}
\begin{equation}
\widehat{\mathbf{a}} = \mathcal{M}(p_{\text{can}}(q, C^{++})). \nonumber
\end{equation}
\end{minipage}
}

We then perform conditional summarization on the retrieved passages \(C^{++}\) by generating summaries \(s_{k}\) that focus on integrating relevant supporting contexts to validate each potential answer \(\widehat{a}_{k} \in \widehat{\mathbf{a}}\) in relation to the question \(q\) as follows:

\vspace{-4mm}
\resizebox{0.95\linewidth}{!}{
\hspace{0mm}\begin{minipage}{\linewidth}
\begin{equation}
s_{k} = \mathcal{M}\left(p_{\text{sum}}\left(q, C^{++}, \widehat{a}_{k}\right)\right) \quad \text{for} \quad k = 1, \ldots, K \nonumber
\end{equation}
\end{minipage}
}

\vspace{1mm}
where \(s_{k}\) represents the conditional summary for the \(k\)-th answer candidate. \(p_{\text{sum}}\) is a prompt designed to facilitate the conditional summarization, extracting the relevant supporting evidence from \(C^{++}\) for the answer candidate \(\widehat{a}_{k}\) in relation to the question \(q\). The probability of generating the conditional summary \(s_{k}\) for the given question \(q\), answer candidate \(\widehat{a}_{k}\), and the retrieved documents \(C^{++}\) can be expressed as:

\vspace{-3mm}
\resizebox{0.925\linewidth}{!}{
\hspace{0mm}\begin{minipage}{\linewidth}
\begin{equation}
P(s_{k} \mid q, \widehat{a}_k, C^{++}) = \prod_{i=1}^{|s_{k}|} P(s_{k_{i}} \mid s_{k_{<i}}, q, \widehat{a}_k, C^{++}) \nonumber
\end{equation}
\end{minipage}
}

where \(s_{k_{i}}\) is the \(i\)-th token of the generated summary \(s_{k}\). \(s_{k_{<i}}\) represents the tokens before \(i\) in the summary. \(|s_{k}|\) denotes the length of the output summary sequence. The core assumption underpinning this approach suggests that focusing on generating strong summaries can lead to more accurate question answering. In simple terms, well-supported and logically valid summaries increase the likelihood of factually correct answers to the question. Subsequently, the multimodal model chooses the corresponding answer from the most plausible summary as the most likely answer to the question through proper evaluation \cite{kim2024sure}. To achieve this evaluation, we utilize a two-pronged approach: (1) Instance-wise validity: This determines whether each generated summary (\(s_{k}\)) is well-formed and provides valid supporting evidence for its corresponding answer candidates (\(\widehat{a}_{k}\)) to the question \(q\). (2) Pair-wise ranking: This compares the generated summaries to determine their relative strength in supporting their respective answer candidates to the question \(q\), identifying the summary that is most plausible and provides the strongest evidence for its corresponding answer candidate. The proposed evaluation approach assesses both the relevance and strength of summaries to select the most plausible answer to a question based on the most compelling and relevant supporting evidence.

\vspace{-6mm}
\subsection{Instance-wise/Pair-wise validity}
\vspace{-2mm}
To evaluate the validity and relevance of each summary \(s_k\) in supporting its corresponding answer candidate \(\widehat{a}_k\), we utilize a two-step validation process. First, we check if \(s_k\) is degenerate, i.e., if it fails to provide meaningful support for \(\widehat{a}_k\) due to insufficient information from the retrieved passages. Second, for non-degenerate summaries, we evaluate how well \(s_k\) specifically supports \(\widehat{a}_k\) compared to other potential answers \(\widehat{a}_i\), where \(i \neq k\). This ensures that the summary \(s_k\) is focused and relevant to the specific answer \(\widehat{a}_k\) it aims to validate. To quantify the validity of each summary, we define a validity score \(v_k\) using a custom prompt \(p_{\text{val}}\), which guides the multimodal model in evaluating the summary's validity \(s_k\) and alignment with the corresponding answer candidate \(\widehat{a}_k\) as follows:

\vspace{-1mm}
\resizebox{0.95\linewidth}{!}{
\hspace{0mm}\begin{minipage}{\linewidth}
\begin{equation}
v(s_k) = 
\begin{cases}
1, & \text{if } \mathcal{M}(p_{\mathrm{val}}(q, \widehat{a}_k, s_k)) = \text{True} \\ \nonumber
0, & \text{otherwise}
\end{cases}
\end{equation}
\end{minipage}
}

The validity score \(v(s_k)\) is 1 if the multimodal model \(\mathcal{M}\), guided by \(p_{\text{val}}\), determines that \(s_k\) is well-formed and supports \(\widehat{a}_k\) in the context of \(q\). Otherwise, \(v(s_k)\) is 0, indicating that \(s_k\) is degenerate or fails to support \(\widehat{a}_k\). To further improve the evaluation of answer plausibility, we introduce a comparative analysis that measures the relative informativeness of each summary in the context of the given question. This evaluation involves comparing a specific summary \(s_k\) to all other generated summaries \(\{s_i\}_{i=1, i \neq k}^{K}\) to determine its effectiveness in providing relevant and valuable information for answering the question \(q\). We adopt a pairwise ranking approach to compute a ranking score \(r\) for each summary \(s_k\), quantifying its performance relative to other summaries. The ranking score is defined as:

\vspace{-5mm}
\resizebox{0.905\linewidth}{!}{
\hspace{0mm}\begin{minipage}{\linewidth}
\begin{equation}
r(s_k, S_K) = \sum_{i \neq k}^{K} r_{\text{pair}}(s_k, s_i) \nonumber
\end{equation}
\end{minipage}
}

\vspace{-1mm}
where \(r_{\text{pair}}(s_k, s_i)\) is the pairwise ranking score between \(s_k\) and \(s_i\), and \(S_K = \{s_1, \ldots, s_K\}\) is the set of all summaries. The term \(r(s_k, S_K)\) represents the overall ranking score for a specific summary \(s_k\) in providing relevant and valuable information for the question \(q\) compared to all other generated summaries. It quantifies the informativeness and relevance of \(s_k\) relative to \(\{s_i\}_{i=1, i \neq k}^{K}\), identifying the summary that best supports the most plausible answer to the question. The pairwise ranking score \(r_{\text{pair}}(s_k, s_i)\) is obtained by leveraging a custom prompt \(p_{\text{rank}}\) that directs the multimodal model \(\mathcal{M}\) to determine which of the two summaries, \(s_k\) or \(s_i\), provides more relevant information for answering the question \(q\). The pairwise ranking score is defined as:

\vspace{-1mm}
\resizebox{0.905\linewidth}{!}{
\hspace{0mm}\begin{minipage}{\linewidth}
\begin{equation}
r_{\text{pair}}(s_k, s_i) = \begin{cases}
1, & \text{if } \mathcal{M}(p_{\text{rank}}(q, s_k, s_i)) = s_k \\ \nonumber
0, & \text{if } \mathcal{M}(p_{\text{rank}}(q, s_k, s_i)) = s_i \\   \nonumber
0.5, & \text{otherwise} \nonumber
\end{cases}
\end{equation} \nonumber
\end{minipage}
\vspace{-1mm}
}

When the model cannot make a clear determination, \(r_{\text{pair}}(s_k, s_i) = 0.5\), indicating equal informativeness. The final answer prediction, \(\widetilde{a}\), is determined by selecting the answer candidate with the highest combined score, considering both the validity score \(v(s_k)\) and the ranking score \(r(s_k, S_K)\) of its corresponding conditional summary \(s_k\):

\vspace{-1mm}
\resizebox{0.925\linewidth}{!}{
\hspace{0mm}\begin{minipage}{\linewidth}
\begin{equation}
\widetilde{a} = \widehat{a}_{k^{\ast}}, \quad k^{\ast} = \underset{k}{\arg \max} \ [v(s_k) + r(s_k, S_K)]   \nonumber
\end{equation}
\end{minipage}
}

The chosen answer (\(\widetilde{a}\)) is supported by a relevant summary (\(s_{k^{\ast}}\)) and is the most informative among the candidates. The prompts (\(p_{\text{can}}, p_{\text{sum}}, p_{\text{val}},\) and \(p_{\text{rank}}\)) are generalizable across datasets and multimodal models. The sub-agent relays the generated answer back to the main agent. However, multimodal models often generate inaccurate or non-truthful responses and struggle to verify and correct their outputs without external feedback. They lack the ability to critically evaluate their responses. To address this, we utilize an iterative verify-then-correct process to refine responses without needing large-scale human annotations.

\vspace{-5mm}
\subsection{Iterative Self-Correction through Reflection}
\vspace{-2mm}
We now enable the main agent to self-correct the sub-agent's outputs based on feedback obtained during verification by a critique agent. The sub-agent's output undergoes progressive refinement through repeated cycles of verification and correction until a predetermined termination criterion is satisfied, such as attaining a target accuracy threshold or executing a fixed number of iterations. The iterative process involves two key steps: reflection and correction. The `reflection' step refers to the verification process, where the main agent relays the generated answer to a critique agent. The critique agent utilizes a high-quality benchmark, like GPT-4 Turbo, to evaluate the generated answer's quality based on question-answer relevance, factual correctness, and comparison to the ground truth using NLP metrics like BLEU, ROUGE, and METEOR. The purpose is to evaluate and provide feedback on the sub-agent's output to determine whether it meets the desired standards of accuracy and truthfulness. The `correction' step occurs after the reflection step. Based on the feedback generated during verification, the main agent delegates the task to the sub-agent to correct and improve its output. In summary, the reflection (verification of the sub-agent's output by the critique agent to generate feedback) and correction cycle (using the feedback from the verification step to revise and improve the previous output) can be repeated iteratively, enabling progressive self-improvement of the sub-agent's generated answer. This process allows the sub-agent to ground its reasoning in factual knowledge, reducing hallucination and leading to more accurate outputs. Additionally, the interleaved reasoning traces provide transparency, resulting in a more interpretable and trustworthy VQA framework. As mentioned earlier, the multi-agent framework consists of (a) an introspective agent, which includes (i) a main agent and (ii) a critique agent. The main agent plays a crucial role by delegating tasks to the relevant sub-agents and coordinating the iterative self-correction through reflection with the critique agent. Given the question (\(q\)) and the retrieved relevant passages (\(C^{++}\)) from parsed documents, memory databases, web articles, and Wikipedia, the sub-agent generates an initial output (\(\widetilde{a}_{i}\)) as follows:

\vspace{-3mm}
\resizebox{0.965\linewidth}{!}{
\hspace{0mm}\begin{minipage}{\linewidth}
\begin{equation}
\widetilde{a}_{i} = \widehat{\mathbf{a}}_{k^{\ast}} = \mathcal{M}(p_{\text{can}}(q, C^{++}))_{k^{\ast}}
\label{eq:1}
\end{equation}
\vspace{-4mm}
\end{minipage}
}

\vspace{-3mm}
\resizebox{0.965\linewidth}{!}{
\hspace{0mm}\begin{minipage}{\linewidth}
\begin{equation}
k^{\ast} = \underset{k}{\arg \max} \ [v(\mathcal{M}(p_{\text{sum}}(q, C^{++}, \widehat{a}_k))) + r(\mathcal{M}(p_{\text{sum}}(q, C^{++}, \widehat{a}_k)), S_K)] \label{eq:2}
\end{equation}
\end{minipage}
}

\vspace{1mm}
Where \(\widehat{a}_k \in \widehat{\mathbf{a}} = \mathcal{M}(p_{\text{can}}(q, C^{++}))\) are the generated answer candidates, \(v(\cdot)\) represents the instance-wise validity score. \(r(\cdot, S_K)\) is the ranking score of a summary \(s_k\), comparing its relevance and informativeness against other generated summaries in the set \(S_K = \{\mathcal{M}(p_{\text{sum}}(q, C^{++}, \widehat{a}_k))\}_{k=1}^K\). \(\widehat{\mathbf{a}}_{k^{\ast}}\) denotes the specific answer candidate from \(\widehat{\mathbf{a}}\) identified as the most plausible or optimal. The index \(k^{\ast}\) is determined by the arg max operation, selecting the index \(k\) that maximizes the combined score of validity and ranking. Equations \eqref{eq:1} and \eqref{eq:2} detail the steps of generating answer candidates, performing conditional summarization, evaluating the validity of each summary, and ranking the summaries to select the most plausible answer candidate \(\widetilde{a}_{i}\) to the question \(q\). Given the output \(\widetilde{a}_{i}\), the main agent delegates it to the critique agent, denoted by \(\mathcal{M}_{Ref}\), to generate feedback \(c_{i}\):

\vspace{-4mm}
\resizebox{0.965\linewidth}{!}{
\hspace{0mm}\begin{minipage}{\linewidth}
\begin{equation}
  c_i \sim \mathcal{M}_{Ref}(q, C^{++}, \widetilde{a}_{i}, \mathcal{T}) \nonumber
\end{equation}
\end{minipage}
}

\vspace{1mm}
where \(c_i\) provides feedback to the main agent. The mathematical function \(\mathcal{T}\) evaluates the sub-agent's answer against GPT-4 Turbo generated gold standards
(ground-truth) for relevance and factual accuracy. The sub-agent corrects the previous output \(\widetilde{a}_{i}\) using the question \(q\), retrieved passages \(C^{++}\), and feedback \(c_i\) to generate an improved output \(\widetilde{a}_{i+1}\) as follows:

\vspace{0mm}
\resizebox{0.965\linewidth}{!}{
\hspace{0mm}\begin{minipage}{\linewidth}
\begin{equation}
    \widetilde{a}_{i+1} = \widehat{\mathbf{a}}_{k^{\ast}} = \mathcal{M}(p^{\prime}_{\text{can}}(q, C^{++}, \widetilde{a}_{i}, c_{i}))_{k^{\ast}}  \nonumber
\end{equation}
\end{minipage}
}

\vspace{1mm}
where \(p^{\prime}_{\text{can}}\) is the updated candidate generation prompt. The updated prompt now includes additional context or feedback \(c_i\) to refine and improve the previous output \(\widetilde{a}_{i}\). In essence, this approach helps in creating a more accurate, improved, and relevant answer, \(\widetilde{a}_{i+1}\), by incorporating the critique agent's feedback. 

\vspace{-5mm}
\subsection{Vision-Language Instruction/Preference Tuning} 
\vspace{-2mm}
We bridge the gap between the general knowledge of pre-trained student models and new task requirements by updating the student model's domain knowledge through vision-language instruction tuning on a task-specific dataset (input-output pairs + instructions), all while mitigating the catastrophic forgetting of pre-trained knowledge. Instruction and preference tuning are necessary to adapt and align student models to particular domain-specific tasks and human preferences, respectively. This typically involves a multi-stage approach: (a) instruction-tuning using task-specific data to adapt the student model to the target task by minimizing the cross-entropy loss, followed by (b) direct preference optimization (DPO) for preference alignment using human preference data to increase the likelihood of generating preferred responses over rejected responses for the target task, thereby minimizing the binary cross-entropy loss. Traditionally, these methods require extensive and often expensive expert-annotated data. In this study, we use teacher-student transfer learning to avoid expensive manual data labeling. Teacher models, trained on vast labeled datasets, transfer their task-specific knowledge to student models through knowledge distillation. This enables student models to achieve high task-specific performance comparable to proprietary teacher models without relying on extensive human annotation. We use parameter-efficient fine-tuning (PEFT) with quantization to adapt pre-trained student models to various tasks by updating a small subset of additional parameters. This reduces memory usage and computational overhead, enabling efficient training and scaling on consumer hardware. We utilize OpenAI GPT-4(Omni) and Google Gemini Pro as teacher models to generate a customized instruction-following dataset of image-question-answer (IQA) triplets and human preference data (image-question-chosen-rejected quadruples). The machine-generated data is tailored to customize student models for image captioning and VQA tasks on PFD and P\&ID analysis. Furthermore, we leverage the Google Cloud Vision API for text detection and OCR tasks, generating IQA triplet data for bounding boxes and detected text recognition tasks. We employ PaliGemma-8K-instruct, a single-turn vision-language model, as the student model. The instruction-tuned student model excels in single-turn analysis but struggles with multi-turn conversations due to its inability to maintain context. We use external databases to store relevant QA pairs and their context. Sub-agents query these databases to retrieve and analyze relevant pairs, enabling them to generate contextually appropriate responses for new, related questions.

\vspace{-5mm}
\section{Experiments}
\vspace{-4mm}
We evaluated our multi-agent framework on open-domain and close-domain QA tasks for analyzing complex PFDs and P\&IDs through image captioning, VQA, and OCR tasks. We built a dataset from academic sources, industry examples, and public repositories, comprising 75 PFDs and 50 P\&IDs. We generated image captions and detailed text descriptions of the PFDs and P\&IDs using GPT-4(Omni). These were compiled into a PDF document, resulting in two distinct categories: \(D_{PFDs}\) containing the PFDs with their captions and descriptions, and \(D_{P\&IDs}\) containing the P\&IDs with their captions and descriptions. PFD and P\&ID documents were parsed using a sliding window technique to improve information retrieval. Text chunks were embedded and indexed, while images were processed using GPT-4 to generate text descriptions, which were then indexed for multi-modal search. OpenAI GPT-4(Omni) and Google Gemini Pro were utilized as teacher models to generate high-quality instruction-tuning and preference-tuning data, including image-question-answer triplets and image-question-chosen-rejected pairs, tailored for PFD and P\&ID analysis. We generated diverse QA pairs to address domain-specific challenges and ensure a high-quality machine-generated dataset, including 625 image captioning QA triplets, 16,000 VQA pairs, and 10,500 text detection and OCR annotations (including image-augmented data). These datasets were split into 70$\%$  training, 15$\%$  validation, and 15$\%$  test sets. They are essential for building and evaluating a robust multi-agent framework capable of handling real-world PFD and P\&ID analysis tasks.
We compared the proposed framework's performance against baseline models, including proprietary models, on image captioning, VQA, text detection, and OCR tasks. The baselines include GPT-4 Turbo-preview, Claude-3 Opus, and Google Gemini 1.0 Pro for a rigorous comparison with state-of-the-art LMMs. For image captioning and VQA (including logical, common sense, and multi-step reasoning), we used BLEU, ROUGE, and METEOR to evaluate caption accuracy versus ground truth. For multiple-choice VQA tasks, we used precision, recall, F1, and exact match to measure answer correctness against ground truth. Evaluating text detection and OCR involved metrics for localization accuracy (Bounding Box Precision, Recall, F1-Score, Intersection over Union (IoU)) and recognition quality (Character Error Rate (CER), Word Error Rate (WER)). The main agent uses Google Gemma-7b-it. The sub-agents use PaliGemma-3b-mix, and the critique agent uses GPT-4 Turbo. We use the open-source BGE embedding method as a search engine to retrieve relevant passages from external sources for knowledge-augmented text generation. We fine-tune an adapter for task-specific customization to improve BGE embedding retrieval performance for PFD and P\&ID analysis. The adapter is trained to rank relevant documents higher than irrelevant passages for a given query by learning the semantic relationships between similar questions and their corresponding answers, enhancing retrieval quality. Similarly, we use the open-source BGE rerank model to prioritize the most relevant and reliable information from retrieved passages. We fine-tune the reranker to assign higher relevance scores to passages that are more relevant to the given queries, ensuring they are ranked higher in the results. We performed instruction-tuning of each SLM (PaliGemma) using the PEFT technique, such as QLoRA, on their specific PFD and P\&ID analysis tasks using corresponding datasets. We resized PFD and P\&ID images to 224$\times$224 for image captioning and VQA tasks or 448$\times$448 for OCR tasks, using bicubic resampling with a patch size of 14$\times$14 pixels. We generated image tokens based on resolution: 256 tokens for 224$\times$224 images and 1024 tokens for 448$\times$448 images. These image tokens were combined with text inputs to PaliGemma for autoregressive text generation.
Each SMM's instruction-tuning leveraged a comprehensive hyperparameter configuration: a batch size of 16, a learning rate of \(1 \times 10^{-3}\) adjusted with a linear scheduler over 40 epochs, 100 warmup steps, a weight decay of \(1 \times 10^{-4}\), gradient accumulation of 5 steps, and the AdamW optimizer. To ensure efficient parameter updates, we utilized 4-bit QLoRA with a low-rank \(r\) of 12, \(\alpha\) of 32, and a dropout of 0.05. We performed preference tuning on each SMM using the DPO technique along with QLoRA, minimizing the binary cross-entropy (BCE) loss with the following hyperparameters: a learning rate of \(5.0 \times 10^{-4}\) with a cosine scheduler and gradient accumulation of 4 steps. \(\beta\) was set to 0.2 to align SLMs with the desired preferences. We conducted training for 20 epochs using the AdamW optimizer, with a batch size of 16. We utilized NVIDIA GPUs for faster training and, for robust evaluation, performed multiple independent runs and reported ensembled averages.

\vspace{-3mm}
\paragraph{\textbf{Experimental Results:}}
The experimental results demonstrated that our framework performed on par with or exceeded state-of-the-art methods in image captioning, VQA, and OCR tasks, while offering customizability, interpretability, data privacy, and cost-effectiveness. A qualitative analysis of the generated outputs highlighted the framework's ability to produce contextually relevant and factually accurate responses for complex PFDs and P\&IDs analysis tasks. Tables \ref{results1}-\ref{results2} show the experimental results for image captioning and VQA tasks. The evaluation metrics BLEU-2, BLEU-4, ROUGE-1, ROUGE-2, ROUGE-L, and METEOR range from 0 (no overlap) to 1 (perfect overlap), with higher scores indicating better performance. Table \ref{results3} shows the experimental results on multiple-choice VQA tasks in terms of precision, recall, F1-score, and exact match(classification task) from 0.0 to 1.0 (higher is better). Table \ref{results4} presents the experimental results on the text detection task using Bounding Box Precision, Recall, F1-Score, and IoU (0.0-1.0 scale, higher is better). Table \ref{results5} shows the experimental results on OCR to recognize and transcribe text in images compared to ground truth text. Lower CER and WER values closer to 0 indicate superior performance. Our framework surpasses or matches top baselines (Tables \ref{results1}-\ref{results5}). To understand the contribution of individual components to overall framework performance, we conducted ablation studies.  We disabled components like instruction-tuning (IT), preference-tuning (PT), iterative self-correction (SC), and conditional summarization (CS) individually to create ablated variants. We compared the performance of the ablated variants with the original(baseline) model. Tables \ref{abaresults1} - \ref{abaresults3} show ablation study results for captioning and VQA, while Tables \ref{abaresults4} - \ref{abaresults5} show results for text detection and OCR.

\vspace{-6mm}
\begin{table*}[!ht]
\centering
\caption{The table compares the proposed method's image captioning performance.}
\vspace{-3mm}
\scalebox{0.785}{
\begin{tabular}{l|c|c|c|c|c}
\toprule
Method & BLEU-2($\uparrow$) & BLEU-4($\uparrow$) & ROUGE-2($\uparrow$) & ROUGE-L($\uparrow$) & METEOR($\uparrow$) \\ 
\midrule
InstructBLIP\cite{dai2305instructblip} & 0.661 $\pm$ 0.054 & 0.620 $\pm$ 0.063 & 0.706 $\pm$ 0.073 & 0.761 $\pm$ 0.026 & 0.792 $\pm$ 0.065 \\ 
\midrule
LLaVA\cite{liu2023visual} & 0.670 $\pm$ 0.062 & 0.623 $\pm$ 0.071 & 0.708 $\pm$ 0.082 & 0.761 $\pm$ 0.034 & 0.789 $\pm$ 0.055 \\ 
\midrule
MiniGPT-4\cite{zhu2023minigpt} & 0.729 $\pm$ 0.101 & 0.637 $\pm$ 0.121 & 0.741 $\pm$ 0.025 & 0.776 $\pm$ 0.086 & 0.808 $\pm$ 0.085 \\ 
\midrule
\textbf{(Ours)} W/GPT-4 Turbo-P & 0.923 $\pm$ 0.074 & 0.904 $\pm$ 0.083 & 0.932 $\pm$ 0.091 & 0.949 $\pm$ 0.038 & 0.946 $\pm$ 0.075 \\ 
\midrule
\textbf{(Ours)} W/Claude-3 Opus & 0.917 $\pm$ 0.081 & 0.896 $\pm$ 0.092 & 0.926 $\pm$ 0.103 & 0.937 $\pm$ 0.045 & 0.932 $\pm$ 0.084 \\ 
\midrule
\textbf{(Ours)} W/Gemini 1.0 Pro & 0.945 $\pm$ 0.105 & 0.928 $\pm$ 0.126 & 0.957 $\pm$ 0.029 & 0.962 $\pm$ 0.086 & 0.943 $\pm$ 0.089 \\ 
\midrule
$\textbf{(Ours) W/PaliGemma}$ & \textbf{0.936 $\pm$ 0.123} & \textbf{0.921 $\pm$ 0.141} & \textbf{0.941 $\pm$ 0.105} & \textbf{0.951 $\pm$ 0.073} & \textbf{0.956 $\pm$ 0.106} \\ 
\bottomrule
\end{tabular}
}
\label{results1}
\vspace{-15mm}
\end{table*}

\vspace{0mm}
\begin{table*}[!ht]
\centering
\caption{The table shows various methods' open/closed-ended VQA performance.}
\vspace{-3mm}
\scalebox{0.80}{
\begin{tabular}{l|c|c|c|c|c}
\toprule
Method                                                                  & BLEU-2 ($\uparrow$)                     & BLEU-4 ($\uparrow$)                     & ROUGE-2 ($\uparrow$)                    & ROUGE-L ($\uparrow$)                   & METEOR ($\uparrow$)                     \\ \midrule
InstructBLIP\cite{dai2305instructblip}  & 0.650$\pm$0.09 & 0.520$\pm$0.11 & 0.660$\pm$0.03 & 0.720$\pm$0.07 & 0.770$\pm$0.09          \\ \midrule
LLaVA\cite{liu2023visual} & 0.658$\pm$0.10 & 0.530$\pm$0.12 & 0.665$\pm$0.04 & 0.720$\pm$0.07 & 0.770$\pm$0.09          \\ \midrule
MiniGPT-4\cite{zhu2023minigpt}  & 0.680$\pm$0.11 & 0.545$\pm$0.13 & 0.675$\pm$0.05 & 0.735$\pm$0.08 & 0.800$\pm$0.10          \\ \midrule
\textbf{(Ours)} W/GPT-4 Turbo-P   & 0.897$\pm$0.10 & 0.871$\pm$0.11 & 0.905$\pm$0.12 & 0.916$\pm$0.06 & 0.929$\pm$0.10          \\ \midrule
\textbf{(Ours)} W/Claude-3 Opus         & 0.883$\pm$0.11 & 0.863$\pm$0.12 & 0.887$\pm$0.13 & 0.895$\pm$0.07 & 0.902$\pm$0.11          \\ \midrule
\textbf{(Ours)} W/Gemini 1.0 Pro    & 0.915$\pm$0.13 & 0.890$\pm$0.14 & 0.924$\pm$0.04 & 0.930$\pm$0.10 & 0.928$\pm$0.11          \\ \midrule
$\textbf{(Ours) W/PaliGemma}$  &  \textbf{0.922$\pm$0.14} & \textbf{0.905$\pm$0.15} & \textbf{0.930$\pm$0.12} & \textbf{0.940$\pm$0.09} & \textbf{0.945$\pm$0.12} \\ 
\bottomrule
\end{tabular}
}
\label{results2}
\vspace{-15mm}
\end{table*}

\vspace{0mm}
\begin{table*}[!ht]
\centering
\caption{The table shows the proposed method's multiple-choice VQA performance.}
\vspace{-3mm}
\scalebox{0.80}{
\begin{tabular}{l|c|c|c|c}
\toprule
Method                                                                  & Precision ($\uparrow$)                     & Recall ($\uparrow$)                     & F1-Score ($\uparrow$)                    & Exact Match ($\uparrow$)                    \\ \midrule
InstructBLIP\cite{dai2305instructblip}  & 0.773$\pm$0.070 & 0.697$\pm$0.094 & 0.890$\pm$0.035 & 0.798$\pm$0.012 \\ \midrule
LLaVA\cite{liu2023visual} & 0.800$\pm$0.078 & 0.720$\pm$0.104 & 0.897$\pm$0.035 & 0.802$\pm$0.012 \\ \midrule
MiniGPT-4\cite{zhu2023minigpt}  & 0.822$\pm$0.084 & 0.735$\pm$0.111 & 0.907$\pm$0.036 & 0.812$\pm$0.013 \\ \midrule
\textbf{(Ours)} W/GPT-4 Turbo-P   & 0.926$\pm$0.095 & 0.906$\pm$0.129 & 0.942$\pm$0.045 & 0.951$\pm$0.015 \\ \midrule
\textbf{(Ours)} W/Claude-3 Opus         & 0.896$\pm$0.088 & 0.884$\pm$0.121 & 0.902$\pm$0.041 & 0.905$\pm$0.014 \\ \midrule
\textbf{(Ours)} W/Gemini 1.0 Pro    & 0.919$\pm$0.095 & 0.892$\pm$0.129 & 0.924$\pm$0.043 & 0.937$\pm$0.015 \\ \midrule
\textbf{(Ours)} W/PaliGemma   & \textbf{0.934$\pm$0.097} & \textbf{0.903$\pm$0.119} & \textbf{0.933$\pm$0.041} & \textbf{0.928$\pm$0.015} \\  \bottomrule
\end{tabular}
}
\label{results3}
\vspace{-7mm}
\end{table*}

\vspace{-7mm}
\begin{table*}[!ht]
\centering
\caption{The table compares our method's text detection accuracy against baselines.}
\vspace{-3mm}
\scalebox{0.825}{
\begin{tabular}{l|c|c|c|c}
\toprule
Method                                                                  & Precision ($\uparrow$)                     & Recall ($\uparrow$)                     & F1-Score ($\uparrow$)                    & IoU ($\uparrow$)                    \\ \midrule
InstructBLIP\cite{dai2305instructblip}   & 0.781$\pm$0.070 & 0.740$\pm$0.087 & 0.827$\pm$0.036 & 0.799$\pm$0.012 \\ \midrule
LLaVA\cite{liu2023visual}         & 0.790$\pm$0.078 & 0.755$\pm$0.097 & 0.839$\pm$0.036 & 0.803$\pm$0.012 \\ \midrule
MiniGPT-4\cite{zhu2023minigpt}    & 0.818$\pm$0.086 & 0.796$\pm$0.101 & 0.825$\pm$0.037 & 0.823$\pm$0.014 \\ \midrule
$\textbf{(Ours) W/PaliGemma}$  &  \textbf{0.907$\pm$0.099} & \textbf{0.873$\pm$0.122} & \textbf{0.915$\pm$0.042} & \textbf{0.879$\pm$0.016} \\ \bottomrule
\end{tabular}
}
\label{results4}
\vspace{-8mm}
\end{table*}

\vspace{-7mm}
\begin{table*}[!ht]
\centering
\caption{The table compares our framework's OCR accuracy to existing methods.}
\vspace{-3mm}
\scalebox{0.825}{
\begin{tabular}{l|c|c|c}
\toprule
Method & CER ($\downarrow$) & WER ($\downarrow$) & F1-Score ($\uparrow$) \\ \midrule
InstructBLIP\cite{dai2305instructblip} & 0.095$\pm$0.070 & 0.130$\pm$0.087 & 0.744$\pm$0.036 \\ \midrule
LLaVA\cite{liu2023visual} & 0.197$\pm$0.078 & 0.230$\pm$0.097 & 0.755$\pm$0.036 \\ \midrule
MiniGPT-4\cite{zhu2023minigpt} & 0.170$\pm$0.086 & 0.191$\pm$0.101 & 0.742$\pm$0.037 \\ \midrule
$\textbf{(Ours) W/PaliGemma}$ & \textbf{0.093$\pm$0.099} & \textbf{0.127$\pm$0.122} & \textbf{0.824$\pm$0.042} \\ \bottomrule
\end{tabular}
}
\label{results5}
\vspace{-9mm}
\end{table*}

A user study evaluated the framework's usability and effectiveness for PFD and P\&ID analysis, gathering feedback on satisfaction and reliability. The findings highlight each component's importance and overall effectiveness.

\vspace{-6mm}
\begin{table*}[!ht]
\centering
\caption{The table shows the ablation study results for the image captioning task.}
\vspace{-3mm}
\scalebox{0.80}{
\begin{tabular}{l|c|c|c|c|c}
\toprule
Method & BLEU-2($\uparrow$) & BLEU-4($\uparrow$) & ROUGE-2($\uparrow$) & ROUGE-L($\uparrow$) & METEOR($\uparrow$) \\ 
\midrule
$\textbf{w/o IT}$ & 0.346 $\pm$ 0.03 & 0.282 $\pm$ 0.03 & 0.372 $\pm$ 0.05 & 0.206 $\pm$ 0.01 & 0.200 $\pm$ 0.04 \\ 
\midrule
$\textbf{w/o PT}$ & 0.816 $\pm$ 0.03 & 0.822 $\pm$ 0.03 & 0.840 $\pm$ 0.05 & 0.848 $\pm$ 0.01 & 0.852 $\pm$ 0.04 \\ 
\midrule
$\textbf{w/o ISC}$ & 0.678 $\pm$ 0.03 & 0.648 $\pm$ 0.03 & 0.744 $\pm$ 0.05 & 0.680 $\pm$ 0.01 & 0.683 $\pm$ 0.04 \\ 
\midrule
$\textbf{w/o CS}$ & 0.688 $\pm$ 0.03 & 0.613 $\pm$ 0.03 & 0.701 $\pm$ 0.05 & 0.629 $\pm$ 0.01 & 0.703 $\pm$ 0.04 \\ 
\midrule
$\textbf{Baseline}$ & \textbf{0.936 $\pm$ 0.123} & \textbf{0.921 $\pm$ 0.141} & \textbf{0.941 $\pm$ 0.105} & \textbf{0.951 $\pm$ 0.073} & \textbf{0.956 $\pm$ 0.106} \\ 
\bottomrule
\end{tabular}
}
\label{abaresults1}
\vspace{-14mm}
\end{table*}

\vspace{0mm}
\begin{table*}[!ht]
\centering
\caption{The table shows the ablation study results on the open-ended VQA task.}
\vspace{-3mm}
\scalebox{0.825}{
\begin{tabular}{l|c|c|c|c|c}
\toprule
Method                                                                  & BLEU-2 ($\uparrow$)                     & BLEU-4 ($\uparrow$)                     & ROUGE-2 ($\uparrow$)                    & ROUGE-L ($\uparrow$)                   & METEOR ($\uparrow$)                     \\ \midrule
$\textbf{w/o IT}$ & 0.197 $\pm$ 0.03 & 0.249 $\pm$ 0.03 & 0.351 $\pm$ 0.05 & 0.220 $\pm$ 0.01 & 0.304 $\pm$ 0.04 \\ 
\midrule
$\textbf{w/o PT}$ & 0.814 $\pm$ 0.03 & 0.798 $\pm$ 0.03 & 0.806 $\pm$ 0.05 & 0.820 $\pm$ 0.01 & 0.822 $\pm$ 0.04 \\ 
\midrule
$\textbf{w/o ISC}$ & 0.687 $\pm$ 0.03 & 0.691 $\pm$ 0.03 & 0.667 $\pm$ 0.05 & 0.718 $\pm$ 0.01 & 0.690 $\pm$ 0.04 \\ 
\midrule
$\textbf{w/o CS}$ & 0.612 $\pm$ 0.03 & 0.643 $\pm$ 0.03 & 0.692 $\pm$ 0.05 & 0.672 $\pm$ 0.01 & 0.654 $\pm$ 0.04 \\ 
\midrule
$\textbf{Baseline}$  &  \textbf{0.922$\pm$0.14} & \textbf{0.905$\pm$0.15} & \textbf{0.930$\pm$0.12} & \textbf{0.940$\pm$0.09} & \textbf{0.945$\pm$0.12} \\ 
\bottomrule
\end{tabular}
}
\label{abaresults2}
\vspace{-14mm}
\end{table*}

\vspace{0mm}
\begin{table*}[!ht]
\centering
\caption{The table presents the ablation study results on multiple-choice VQA tasks.}
\vspace{-3mm}
\scalebox{0.825}{
\begin{tabular}{l|c|c|c|c}
\toprule
Method                                                                  & Precision ($\uparrow$)                     & Recall ($\uparrow$)                     & F1-Score ($\uparrow$)                    & Exact Match ($\uparrow$)                    \\ \midrule
$\textbf{w/o IT}$  &  \textbf{0.274$\pm$0.085} & \textbf{0.236$\pm$0.105} & \textbf{0.286$\pm$0.036} & \textbf{0.267$\pm$0.014} \\ \midrule
$\textbf{w/o PT}$  &  \textbf{0.803$\pm$0.085} & \textbf{0.799$\pm$0.105} & \textbf{0.815$\pm$0.036} & \textbf{0.791$\pm$0.014} \\ \midrule
$\textbf{w/o ISC}$  &  \textbf{0.740$\pm$0.085} & \textbf{0.693$\pm$0.105} & \textbf{0.733$\pm$0.036} & \textbf{0.717$\pm$0.014} \\ \midrule
$\textbf{w/o CS}$  &  \textbf{0.684$\pm$0.085} & \textbf{0.591$\pm$0.105} & \textbf{0.630$\pm$0.036} & \textbf{0.663$\pm$0.014} \\ \midrule
$\textbf{Baseline}$   & \textbf{0.934$\pm$0.097} & \textbf{0.903$\pm$0.119} & \textbf{0.933$\pm$0.041} & \textbf{0.928$\pm$0.015} \\  \bottomrule
\end{tabular}
}
\label{abaresults3}
\vspace{-10mm}
\end{table*}

\vspace{-5mm}
\begin{table*}[!ht]
\centering
\caption{The table shows the ablation study results on the text detection task.}
\vspace{-3mm}
\scalebox{0.825}{
\begin{tabular}{l|c|c|c|c}
\toprule
Method & Precision ($\uparrow$) & Recall ($\uparrow$) & F1-Score ($\uparrow$) & IoU ($\uparrow$) \\ \midrule
$\textbf{w/o IT}$ & 0.127$\pm$0.063 & 0.121$\pm$0.078 & 0.095$\pm$0.032 & 0.108$\pm$0.011 \\ \midrule
$\textbf{w/o PT}$ & 0.774$\pm$0.063 & 0.761$\pm$0.078 & 0.802$\pm$0.032 & 0.745$\pm$0.011 \\ \midrule
$\textbf{w/o ISC}$ & 0.590$\pm$0.063 & 0.530$\pm$0.078 & 0.584$\pm$0.032 & 0.593$\pm$0.011 \\ \midrule
$\textbf{w/o CS}$ & 0.737$\pm$0.063 & 0.731$\pm$0.078 & 0.735$\pm$0.032 & 0.705$\pm$0.011 \\ \midrule
$\textbf{Baseline}$ & \textbf{0.907$\pm$0.099} & \textbf{0.873$\pm$0.122} & \textbf{0.915$\pm$0.042} & \textbf{0.879$\pm$0.016} \\ \bottomrule
\end{tabular}
}
\label{abaresults4}
\vspace{-14mm}
\end{table*}

\vspace{0mm}
\begin{table*}[!ht]
\centering
\caption{The table presents the ablation study results on OCR task.}
\vspace{-3mm}
\scalebox{0.825}{
\begin{tabular}{l|c|c|c}
\toprule
Method & CER ($\downarrow$) & WER ($\downarrow$) & F1-Score ($\uparrow$) \\ \midrule
$\textbf{w/o IT}$ & 0.733$\pm$0.063 & 0.840$\pm$0.078 & 0.098$\pm$0.032 \\ \midrule
$\textbf{w/o PT}$ & 0.105$\pm$0.063 & 0.145$\pm$0.078 & 0.731$\pm$0.032 \\ \midrule
$\textbf{w/o ISC}$ & 0.324$\pm$0.063 & 0.343$\pm$0.078 & 0.542$\pm$0.032 \\ \midrule
$\textbf{w/o CS}$ & 0.208$\pm$0.063 & 0.238$\pm$0.078 & 0.694$\pm$0.032 \\ \midrule
$\textbf{Baseline}$ & \textbf{0.093$\pm$0.099} & \textbf{0.127$\pm$0.122} & \textbf{0.824$\pm$0.042} \\ \bottomrule
\end{tabular}
}
\label{abaresults5}
\vspace{-6mm}
\end{table*}

\begin{figure}
\centering
\resizebox{0.675\linewidth}{!}{ 
\hspace*{0mm}\includegraphics[keepaspectratio,trim=1.5cm 0cm 0cm 0.75cm,clip]{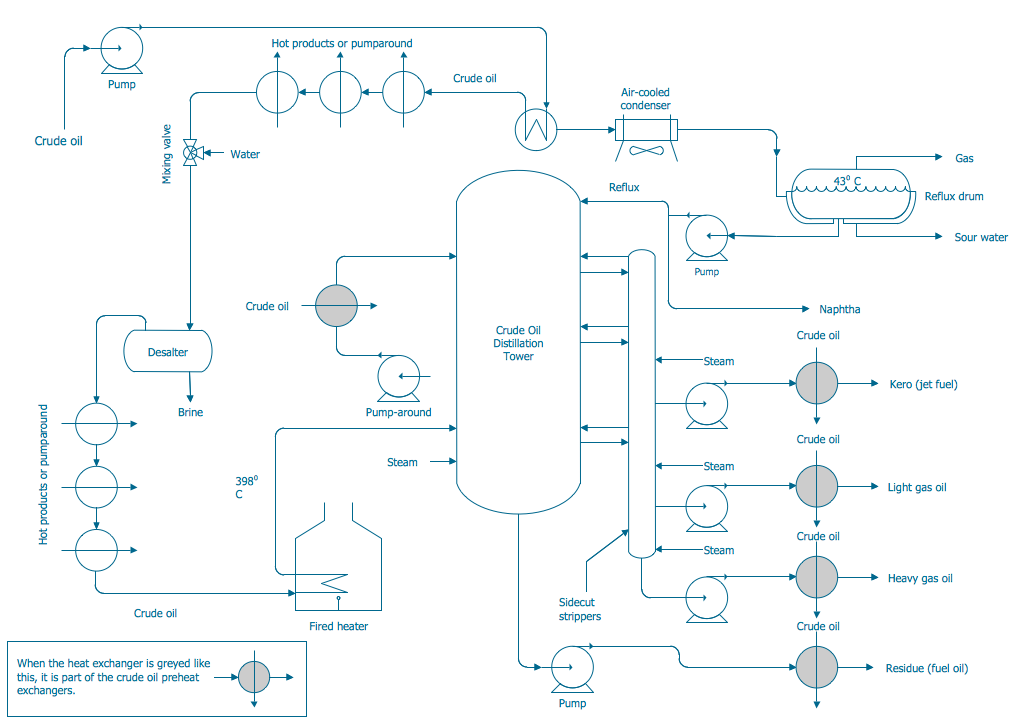} 
}
\vspace{-3mm}
\caption{The figure shows the PFD of a crude oil distillation unit. }
\label{fig:PFDfigure}
\vspace{-3mm}
\end{figure}

\begin{figure}
\centering
\resizebox{0.675\linewidth}{!}{ 
\hspace*{0mm}\includegraphics[keepaspectratio,trim=0.0cm 0cm 0cm 0.5cm,clip]{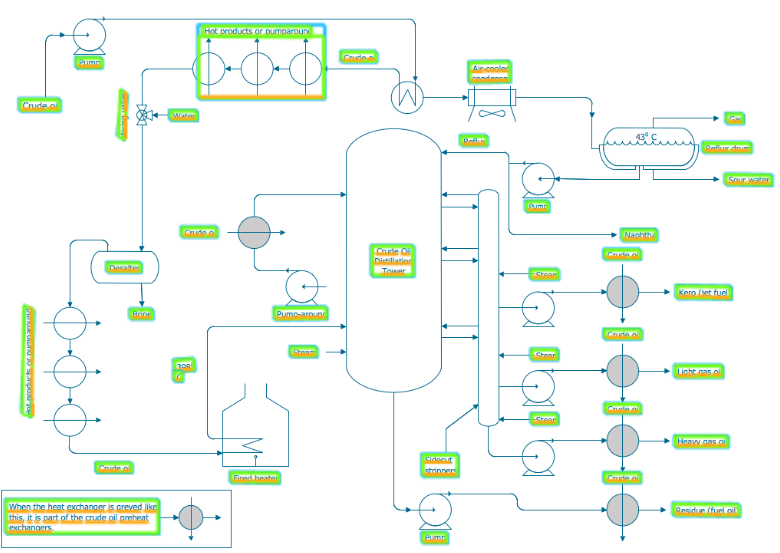} 
}
\vspace{-4mm}
\caption{The figure shows the text detection and OCR results for the PFD.}
\label{fig:textdetectionOCRfigure}
\vspace{-15mm}
\end{figure}

Figures~\ref{fig:PFDfigure} and~\ref{fig:textdetectionOCRfigure} illustrate representative examples of a sample PDF and its corresponding OCR/text detection results. Table~\ref{tab:ansgen} presents illustrative results for image captioning, multiple-choice question answering, and closed- and open-ended visual question answering.

\vspace{-11mm}

\vspace{0mm}
\begin{table*}[ht!]
\begin{tcolorbox}[colback=white, colframe=black, coltitle=black, colbacktitle=white, title=Image Captioning - Multiple-Choice QA - Open/Close-Ended VQA.]
\footnotesize
\vspace{-2mm}
\begin{tcolorbox}[colback=white, colframe=black, coltitle=black, colbacktitle=white, title=Write a high-level figure caption for the provided PFD?.] 
\vspace{-2mm}
\textbf{Answer:} The PFD diagram illustrates the key components and flow paths involved in the distillation process of crude oil. Crude oil is first pumped and heated through a series of exchangers and a desalter before entering the distillation tower. In the tower, the crude oil is separated into various fractions, such as naphtha, kerosene, light gas oil, heavy gas oil, and residue, based on their boiling points. The diagram also shows the use of steam in sidecut strippers and the management of reflux and sour water in the system.
\vspace{-3mm}
\end{tcolorbox}
\vspace{-2mm}
\begin{tcolorbox}[colback=white, colframe=black, coltitle=black, colbacktitle=white, title=How is the heavy gas oil fraction processed in the flow sheet?.]
\footnotesize
\vspace{-1mm}
\begin{enumerate}
\vspace{-1mm}
  \item[A)] It is sent to the reflux drum
  \item[B)] It is collected at the bottom of the distillation tower
  \item[C)] It is stripped using steam
  \item[D)] It is removed as residue (fuel oil)
  \vspace{-3mm}
\end{enumerate}
\textbf{Answer:} C) It is stripped using steam.
\vspace{-3mm}
\end{tcolorbox}
\vspace{-2mm}
\begin{tcolorbox}[colback=white, colframe=black, coltitle=black, colbacktitle=white, title=Explain the role of the reflux drum in the crude oil distillation process?.]
\vspace{-1mm}
\textbf{Answer:} The reflux drum condenses the overhead vapor from the distillation tower. The liquid is partially returned as reflux to the tower, while the rest, along with non-condensable gases, is sent for further processing or storage.
\vspace{-3mm}
\end{tcolorbox}
\vspace{-1mm}
\begin{tcolorbox}[colback=white, colframe=black, coltitle=black, colbacktitle=white, title=Describe the flow path of crude oil from the initial entry to the desalter?.]
\vspace{-1mm}
\textbf{Answer:} Crude oil is pumped from storage and preheated through heat exchangers using hot products or pump-around streams. It then enters the desalter, where water washes out salts and impurities, resulting in desalted crude oil and brine.
\vspace{-2mm}
\end{tcolorbox}
\vspace{-4mm}
\end{tcolorbox}
\vspace{-1mm}
\caption{The table shows illustrative examples of responses for image captioning, multiple-choice QA, and open/close-ended VQA.}
\label{tab:ansgen}
\vspace{-7mm}
\end{table*}

\clearpage
\newpage

\vspace{-8mm}

\begin{table*}[ht!]
\begin{tcolorbox}[colback=white, colframe=black, coltitle=black, colbacktitle=white, title=Prompt for potential answer candidates generation($p_{can}$).]
\footnotesize
\vspace{-1mm}
You will be given several passages relevant to the question. Identify two likely correct answers. 
Format your answers concisely as follows: (a) Answer 1, (b) Answer 2. Keep responses clear and succinct. \\ 
\textbf{Question:} (Question) \\
\textbf{Answer:} 
\end{tcolorbox}
\vspace{-1mm}
\caption{Given a question and relevant passages, prompt a small-scale multimodal model(SMM) to generate multiple potential high-quality answers.}
\label{tab:ansgen_prompt}
\vspace{-19mm}
\end{table*}

\begin{table*}[ht!]
\begin{tcolorbox}[colback=white, colframe=black, coltitle=black, colbacktitle=white,title=Prompt for conditional summarization($p_{\text{sum}}$).]
\footnotesize
\vspace{-1mm}
In this task, act as a process engineer. Craft a high-quality passage that strengthens the given prediction using only the provided supporting passages. \\ 
\textbf{Question:} (Question) \\ 
\textbf{Choices( potential answer candidates):} (a) Choice 1 (b) Choice 2 \\
\textbf{Prediction(Answer candidate that the summary is intended to support:} (a) Choice 1 (or (b) Choice 2) \\
\textbf{Passage(Summary generated to validate and support this prediction as the correct answer):} 
\vspace{-2mm}
\end{tcolorbox}
\vspace{-1mm}
\caption{The conditional prompt instructs the language model to generate summaries supporting each answer candidate based on the information from the retrieved passages.}
\label{tab:cond_prompt}
\vspace{-12mm}
\end{table*}

We use the prompt $p_{can}$ with the question and $N$ passages, generating $K$ answer candidates ${\hat{a}} = {\hat{a}_1, ..., \hat{a}_K}$. For $K=2$, see Table \ref{tab:ansgen_prompt} for details. The prompt $p_{\text{sum}}$ generates conditional summaries $s_k$ that provide explicit rationales extracted from the relevant information in the retrieved passages to support each answer candidate. These summaries help assess if each answer $\widehat{a}_k$ is valid. See Table \ref{tab:cond_prompt} for prompt details. The \(p_{val}\) prompt (refer Table \ref{tab:instval_prompt}) assesses the validity of each generated summary \(s_{k}\). It determines if a summary lacks justification due to insufficient source passages  (non-degenerate \(s_k\)) or if the generated summary \(s_k\)  strongly support \(\widehat{a}_k\) over other potential answers like \(\widehat{a}_j\), where \(j \neq k\). The \(p_{rank}\) prompt (refer Table \ref{tab:pairwise_prompt}) compares summaries to determine which is more informative and relevant, helping select the most plausible answer.

\begin{table*}[ht!]
\vspace{-3mm}
\begin{tcolorbox}[colback=white, colframe=black, coltitle=black, colbacktitle=white, title=Prompt for instance-wise validation($p_{val}$).]
\vspace{-2mm}
Given the following question, answer candidate, and summary, determine if the summary supports the answer candidate based on the provided passages. \\
\textbf{Question}: (Question) \\
\textbf{Answer Candidate}: (Answer Candidate) \\
\textbf{Summary}: (Generated Summary) \\
Is the summary valid? (True/False) 
\vspace{-3mm}
\end{tcolorbox}
\vspace{-1mm}
\caption{The instance-wise validation prompt checks if the summary accurately and logically supports the answer candidate in answering the question.}
\label{tab:instval_prompt}
\vspace{-6mm}
\end{table*}

\begin{table*}[ht!]
\vspace{-1mm}
\begin{tcolorbox}[colback=white, colframe=black, coltitle=black, colbacktitle=white,title=Prompt for pair-wise ranking($p_{rank}$).]
\vspace{-2mm}
Given the following question and two summaries, determine which summary better supports the answer to the question based on the provided passages. \\
\textbf{Question}: (Question) \\
\textbf{Summary 1}: (Summary 1) ; \textbf{Summary 2}: (Summary 2) \\
Which summary is better? (1/2/0)
\vspace{-3mm}
\end{tcolorbox}
\vspace{-1mm}
\caption{The pair-wise ranking prompt compares two summaries to determine which better supports the answer, responding with "1" for Summary 1, "2" for Summary 2, or "0" if equally informative.}
\label{tab:pairwise_prompt}
\vspace{-8mm}
\end{table*}

\vspace{-5mm}
\section{Conclusion}
\vspace{-3mm}
The proposed multi-agent framework significantly advances human-level understanding of complex engineering diagrams, ensuring enhanced data privacy, explainability, and cost-effectiveness while achieving superior performance in PFD and P\&ID analysis. Experimental results confirm the framework's effectiveness, highlighting its transformative potential in the chemical and process industries.

\vspace{-4mm}
\bibliographystyle{splncs04}
\bibliography{mybibliography}

\end{document}